# ENTRENAMIENTO DE UNA RED NEURONAL PARA EL RECONOCIMIENTO DE IMÁGENES DE LENGUA DE SEÑAS CAPTURADAS CON SENSORES DE PROFUNDIDAD

*Rivas P. Pedro E.*[1], *Velarde-Anaya Omar*[1], *González-López Samuel*[1], *Rivas P. Pablo*[2], *Álvarez-Torres Norma Angélica.*[3]

[1]Instituto Tecnológico de Nogales, DEPI, Nogales, Sonora.
[2]Marist College, School of Computer Science and Mathematics, Poughkeepsie, NY.
[3]Universidad Tecnológica de Nogales, Depto. de Mecatrónica, Nogales, Sonora.

ezequiel.rivas@depiitn.edu.mx

## RESUMEN

**Debido al crecimiento de la población con problemas auditivos, se han venido desarrollando dispositivos que facilitan la inclusión de las personas sordas en la sociedad, utilizando la tecnología como herramienta de comunicación, tal como los sistemas de visión. A continuación, se presenta una solución a este problema utilizando redes neuronales y autocodificadores para la clasificación de imágenes de Lenguaje de Señas Americano. Como resultado se obtuvo 99.94% de exactitud y un error del 0.01684 para la clasificación de imágenes.**

## ABSTRACT

**Due to the growth of the population with hearing problems, devices have been developed that facilitate the inclusion of deaf people in society, using technology as a communication tool, such as vision systems. Then, a solution to this problem is presented using neural networks and autoencoders for the classification of American Sign Language images. As a result, 99.5% accuracy and an error of 0.01684 were obtained for image classification.**

## 1. INTRODUCCIÓN

Según datos de INEGI en el 2010 la población [1] con limitación auditiva es del 0.44% del total en México. Según Humphries [2] la pérdida de audición neurosensorial globalmente es un defecto más común en el nacimiento, que ocurre en 2 o 3 de cada 1000 nacimientos en países desarrollados y es mucho más alto en países subdesarrollados, tal como en Nigeria, donde se encuentra este problema en 28 por cada 1000. En las causas pos-natales de pérdida auditiva neurosensorial aumenta la cifra, para la edad escolar, 6 o 7 de cada 1000 niños tienen pérdida auditiva permanente. De acuerdo a investigaciones [2], al 80% de los niños sordos de nacimiento, de la población mundial en países desarrollados, se les implementan dispositivos cocleares que les permiten a algunos de ellos la entrada de sonidos en su temprana edad, lo cual les ayuda a desarrollar el habla. Sin embargo, estos dispositivos pueden llegar a tener un costo muy elevado.
El número de niños sordos que son afectados es bastante grande y de especial cuidado, por lo que se tiene la necesidad de investigar continuamente soluciones para incrementar su calidad de vida.

En este sentido, se han realizado diversas investigaciones al respecto, Oikonomidis [3] presenta una solución novedosa al problema de la recuperación y seguimiento de la posición en 3D, la orientación y la articulación completa de una mano humana a partir de observaciones visuales sin marcas obtenidas por un sensor Kinect. Se trata de un problema de optimización, que busca los parámetros del modelo de mano que reduzcan al mínimo la discrepancia entre la apariencia y la estructura 3D de casos hipotéticos de un modelo de mano y las observaciones de la mano real. El problema se resuelve usando una variante de algoritmo de optimización por enjambre de partículas (PSO) [4]. Esto da como resultado un método robusto y eficiente para el seguimiento de la postura completa de una mano en la compleja articulación. Los resultados experimentales extensivos demostraron que para el seguimiento de mano en 3D se comporta de una forma precisa y robusta a una frecuencia de 15 Hz. Por lo tanto, la evidencia comprueba que es casi factible realizarlo en tiempo real.
En el mismo año, se desarrolló una investigación sobre Kinect [5]. Éste se encuentra sobre un vehículo que se está moviendo en una superficie verde, buscando una bola roja. El Kinect capta las imágenes RGB y de profundidad, las cuales se envían a una tableta mediante un puerto USB. Los resultados obtenidos demostraron que no es completamente eficiente la navegación del robot móvil usando solo visión estéreo, ya que las condiciones del entorno (luminosidad, características de color del objeto, entre otros) hacen que el algoritmo de navegación deba ser ajustado constantemente.

La disponibilidad en el mercado de sensores RGBD fue adoptado con un enfoque de desarrollo prometedor que permitiera construir soluciones confiables y rentables [6]. Usando sensores de profundidad, como Microsoft Kinect u otros dispositivos similares, es posible diseñar reconocimiento de actividad o mapas profundos, los cuales son una buena fuente de información porque no son afectados por la variación de la luz del ambiente. Este sensor, puede proporcionar imágenes como la figura del cuerpo, y simplificar los problemas de detección y segmentación del humano [7].

## 2. METODOLOGIA

### 2.1 Conjunto de datos

Se tiene una colección de 5000 imágenes de profundidad obtenidas del sensor Creative Senz3D a una resolución de 256 x 256, proporcionadas en el repositorio llamado *Fingerspelling Recognition* [8]. La colección consta de 5 sujetos y cada uno realizó 5 señas. Cada sujeto generó 200 imágenes por cada seña, lo cual da un total de 1000 imágenes por sujeto. En la Fig.1 se observan imágenes de cada sujeto con diferente seña (dígitos). El motivo por el cual se escogió esta colección de imágenes, es debido a que, son imágenes capturadas por un sensor de profundidad y sobre todo, son imágenes del Lenguaje de Señas Americano (LSA). Las propiedades que se extraen de estas imágenes son el valor de cada pixel (rango de 0 a 255) dentro de un vector de 256 x 256 que representa cada imagen.

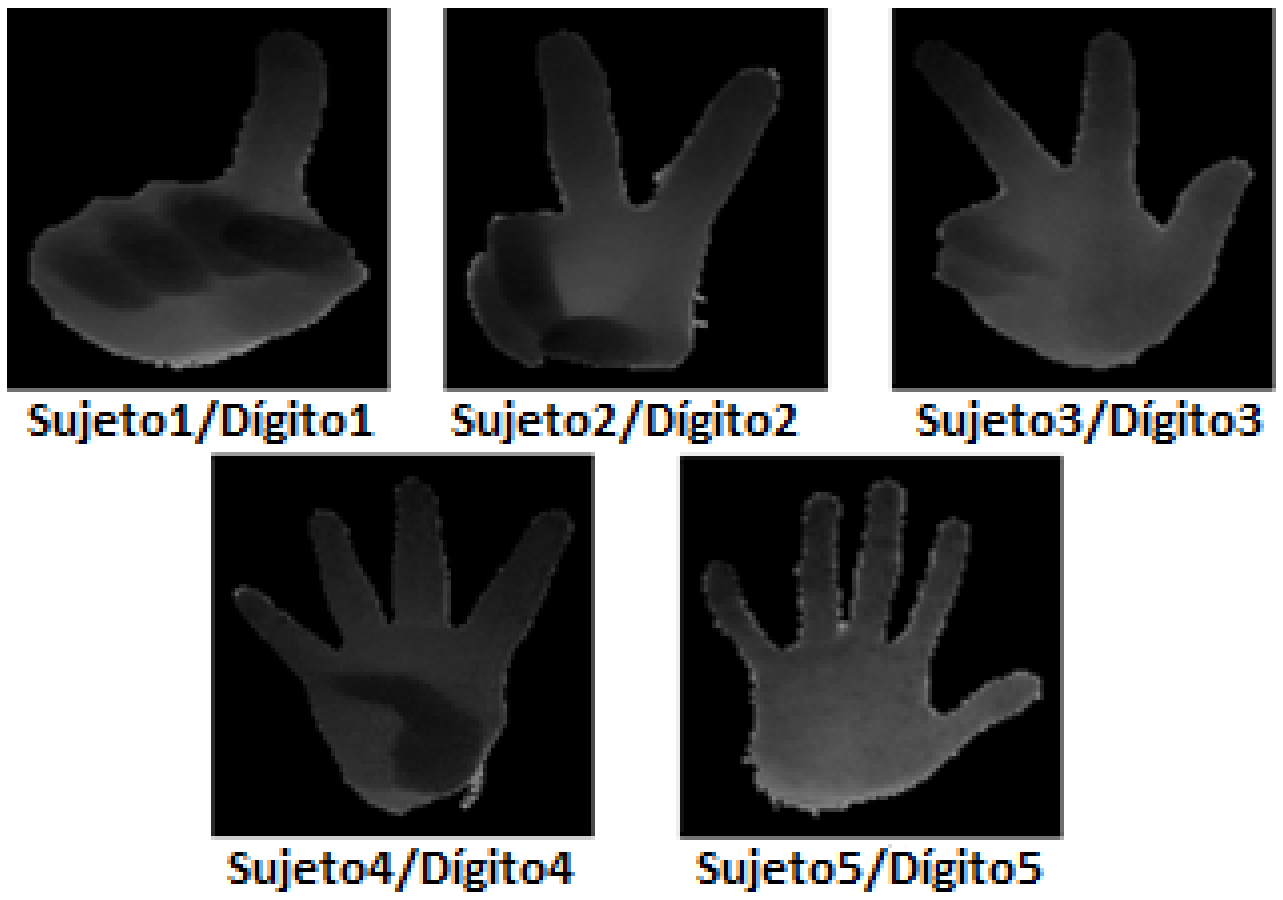

*Fig.1 Imágenes de Lenguaje de Señas por cada sujeto*

### 2.2 Redes Neuronales (RN)

Las redes neuronales son una implementación muy sencilla de un comportamiento local, observado en nuestro cerebro. El cerebro está compuesto de neuronas, las cuales son elementos individuales de procesamiento. La información viaja entre las neuronas, y basado en la arquitectura y ganancia de los conectores neuronales, la red se comporta de forma diferente [9]. Cada neurona está conectada con otra neurona por medio de un peso de ajuste representado por la letra ***w***. Una neurona artificial (Fig. 2) está formada por un sumador y una función de activación, representada por *f(x).*

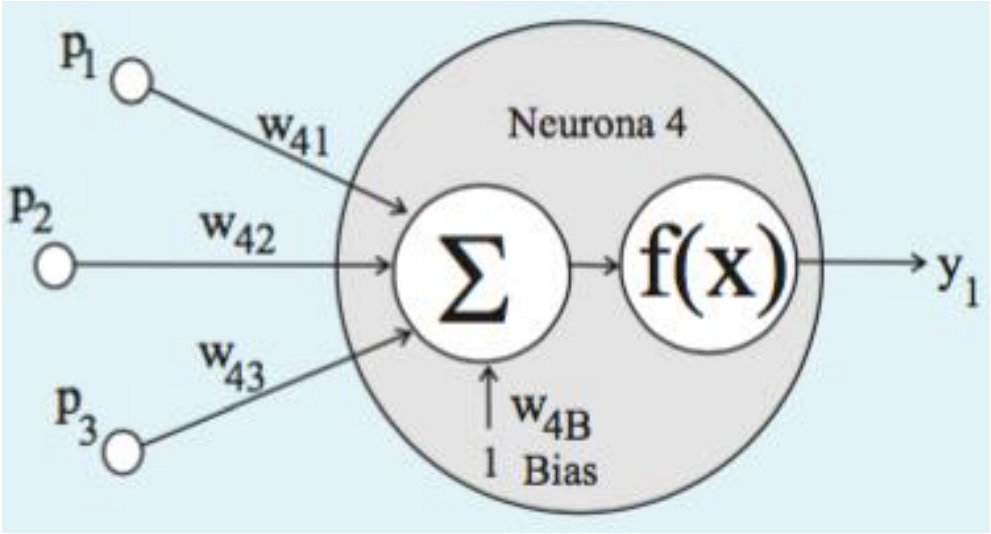

*Fig. 2. Neurona artificial*

Una de las funciones de activación que se utiliza para la clasificación de imágenes es la denominada *softmax*, que es la extensión de la función *sigmoide* para más de dos clases (multiclase). Su papel es de importancia central en muchos modelos probabilísticos no lineales. En particular, en muchos modelos bien conocidos que tratan con datos discretos y continuos [10].

La ecuación de la función *softmax* está dada por [11]:

$$g(Z_m) = \frac{e^{Z_m}}{\sum_k e^{Z_k}} \qquad \text{EC. 1}$$

Para asegurar que la probabilidad de distribución sea válida, $e^{Z_m} > 0$ donde *m* representa las imágenes y $\sum_k e^{Z_k} = 1$.

En este trabajo se manejó la RN simple de una capa con una función de activación *softmax*. Esta función de activación es usada para resolver problemas de función de transferencia no lineal (exponencial) en lugar de logísticas, normalizando las salidas para que su suma sea 1 [12]. Se utilizó esta red con 5 salidas que corresponde a cada seña.

### 2.3 Autocodificadores (AE)

Un AE es una RN que se entrena para intentar copiar su entrada a su salida. Tiene un algoritmo de aprendizaje no supervisado que aplica retro-propagación, estableciendo valores objetivos iguales a los de entrada. El AE intenta aprender una aproximación a la función de identidad, de modo que la salida ***r*** es similar a ***x*** [13]. Internamente, tiene capas ocultas ***h*** que describen el código utilizado para representar la entrada. La red puede ser vista en dos partes: una función *codificador* $h = f(x)$ y un *decodificador* que produce una reconstrucción $r = g(h)$. La arquitectura es representada por la Fig.2. Los AEs podrían ser tomados como un caso especial de redes de realimentación *(feed-forward)*, y podría ser entrenado de igual manera con todas las técnicas, típicamente un gradiente descendiente miniatura, siguiendo el gradiente calculado por la propagación hacia atrás (*back-propagation)*.

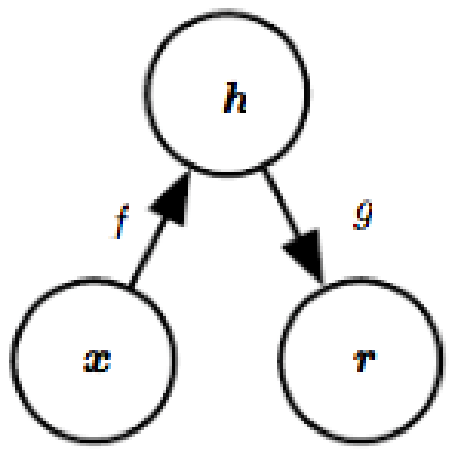

*Fig. 3. La estructura general de un autoencoder, mapeando una entrada **x** a un salida (llamado reconstrucción) **r** a través de una representación interna o codificado **h**. El AE tiene dos componentes: el codificador **f** (mapeo de **x** a **h**) y el decodificador **g** (mapeo de **h** a **r**).*

El mapeo aprendido de la parte del *codificador* de un AE puede ser muy útil para extraer características de la imagen. Cada neurona en la sección *codificadora* tiene un vector de pesos asociado al cual será configurado (*afinación*) para responder a una característica visual en particular. Entrenar el AE usando señales produce resultados como se muestra en la Fig.4 en la capa escondida.

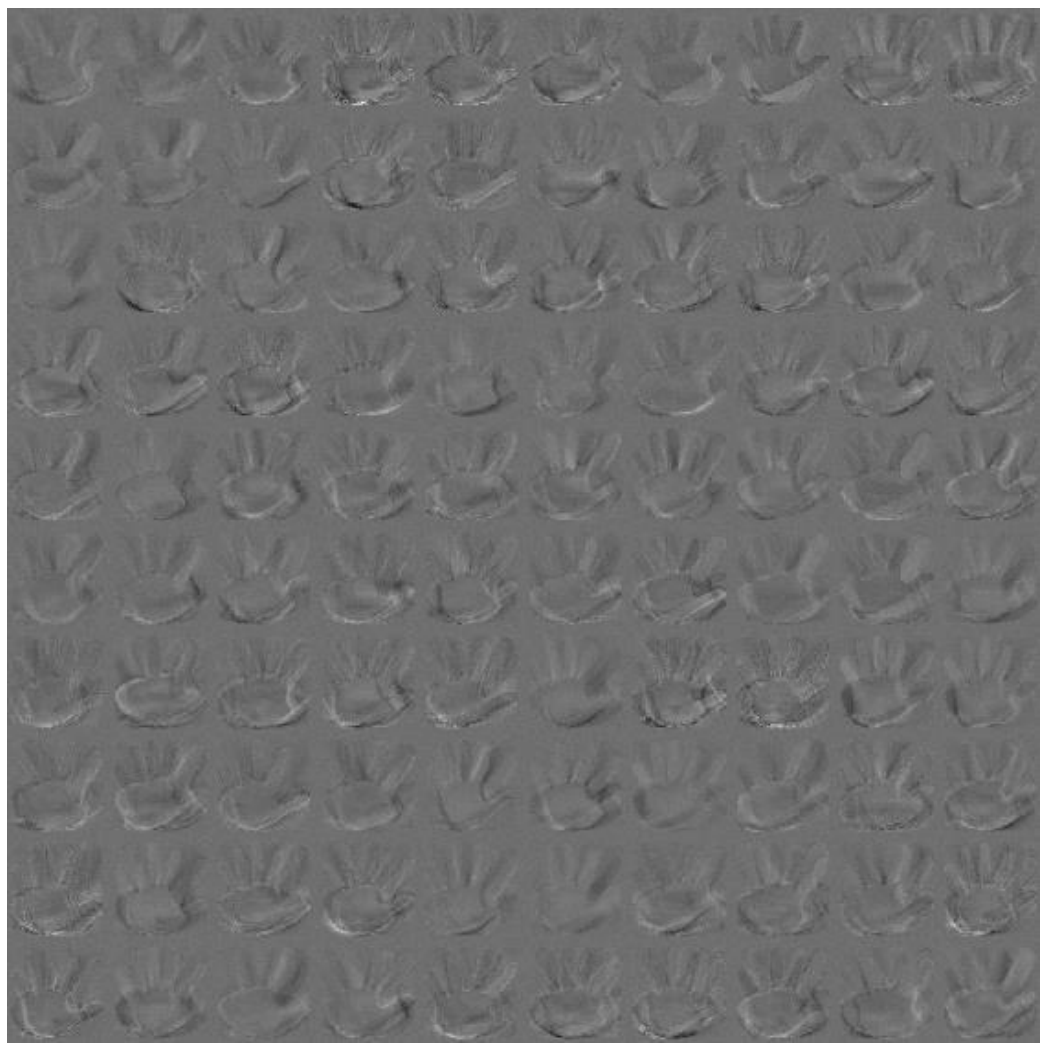

*Fig. 4. Representación de los pesos de cada neurona*

Como se puede observar, los AEs son capaces de capturar características importantes y discriminantes de las imágenes.

Las RNs son populares en tareas de reconocimiento de patrones de imágenes [14]. Han sido de gran interés para la comunidad científica, ya que emular el principal sentido de los seres humanos (la vista) es de suma importancia para poder realizar ciertas tareas automáticamente que en la actualidad requieren un sistema de visión, y que son actividades muy repetitivas, por ejemplo, la verificación de firmas, reconocimiento de caras, reconocimiento de objetos [15], etc. Generalmente se aplican con otros métodos de extracción de características.

## 2.4 Aprendizaje Profundo (AP)

El AP permite que los modelos computacionales que se componen de varias capas de procesamiento, aprendan representaciones de datos con múltiples niveles de abstracción [16]. Estos métodos han mejorado rápidamente el estado de la técnica en reconocimiento de voz, reconocimiento de objetos visuales, detección de objetos y muchos otros dominios como el descubrimiento de fármacos y la genómica.

## 3. EXPERIMENTOS

Se distinguen dos grupos de datos principales, el de entrenamiento y el de prueba. La primera parte como conjunto de entrenamiento se utiliza para determinar los parámetros del clasificador y el conjunto de prueba se utiliza para estimar el error de generalización lo más cercano a cero. El conjunto de entrenamiento suele dividirse en entrenamiento y validación para ajustar el modelo [17].
En base a lo anterior, se utilizó el 50% de los datos para entrenamiento, el 25% para validación y el otro 25% del conjunto de datos para prueba.

Se inició con el aprendizaje de 1000 imágenes (200 por cada seña) del primer sujeto. Después, el segundo sujeto con otras 1000 imágenes, y así consecutivamente hasta llegar al quinto sujeto.

El modelo que se utilizó para el entrenamiento es el AE disperso. En la primera capa maneja regularizadores para aprender una representación dispersa. Algunos parámetros de configuración de la red son:

| | |
|---|---|
| *Entrada*: | 256x256x500, |
| *Capas Ocultas*: | 100, |
| *Cantidad Max. De Épocas*: | 400 |

Después del entrenamiento del primer AE, se entrenó el segundo de una manera similar. La principal diferencia es que se utilizan características del primer AE. También, se disminuyó el tamaño de la representación de la capa oculta.

| | |
|---|---|
| *Entrada*: | 100x500, |
| *Capas Ocultas*: | 50, |
| *Cantidad Max. De Épocas*: | 100 |

Por último, se entrenó la capa *softmax* para clasificar 5 imágenes,

| | |
|---|---|
| *Entrada*: | 50x500, |
| *Capa de Salida*: | 5x500, |
| *Cantidad Max. De Épocas*: | 400 |

Como fue explicado anteriormente, los *codificadores* de los AEs se usaron para extraer características de las imágenes muy representativas. Se pueden apilar AEs junto con la capa de salida *softmax* para formar una RN de aprendizaje profundo (Fig. 6).

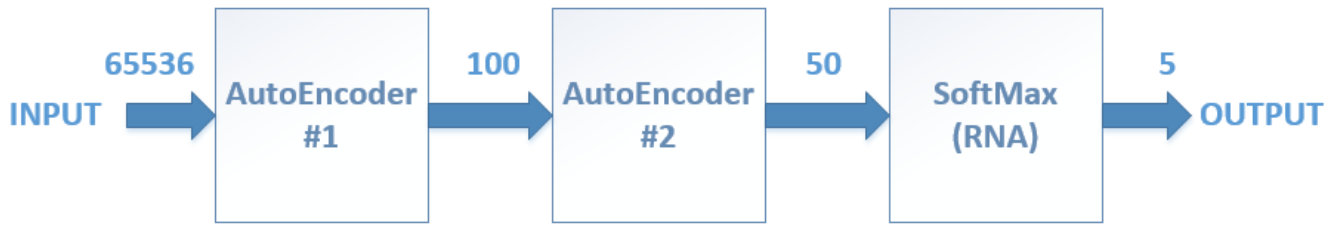

*Fig. 6. Arquitectura de la RN de AP*

Uno de los parámetros [18] para medir el buen rendimiento en regresión es el siguiente:

Error de raíz cuadrático medio normalizado (NRMSE), donde $y_i$ es un vector de *n* predicciones y $d_i$ es un vector de valores verdaderos.

$$NRMSE = \frac{1}{\sigma}\sqrt{\frac{1}{N}\sum_{i=1}^{N}(y_i - d_i)^2} \qquad \text{EC. 2}$$

La matriz de confusión es muy útil en la estimación de los métricos de rendimiento usados en la investigación y son definidos de la siguiente manera:

Exactitud (ACC),

$$ACC = \frac{TP+TN}{TP+FN+FP+TN'} \qquad \text{EC. 3}$$

Tasa de error balanceado (BER),

$$BER = \frac{1}{2}\left(\frac{FP}{TN+FP} + \frac{FN}{FN+TP}\right) \qquad \text{EC. 4}$$

Precisión y exhaustividad (F1-Score),

$$F_1 - Score = 2 \times \frac{\left(\frac{TP}{TP+FP}\right)\times\left(\frac{TP}{TP+FN}\right)}{\left(\frac{TP}{TP+FP}\right)+\left(\frac{TP}{TP+FN}\right)} \qquad \text{EC. 5}$$

Para las ecuaciones anteriores se utilizó la siguiente terminología:
TP (*true positive*) verdadero positivo. Son los casos en el que predice si, y corresponde a la seña.
TN (*true negative*) verdadero negativo. Predecimos un no, y no corresponde a la seña.
FP (*false positive*) falso positivo. Predecimos un si, pero no es la seña.
FN (*false negative*) falso negativo. Predecimos un no, y corresponde a una seña.

## 4. RESULTADOS Y DISCUSIÓN

A continuación, se muestran los resultados del desempeño que se tuvo durante el entrenamiento de la RN para la clasificación de las señas (Tabla 1). Se puede observar que el error (NRMSE) es muy cercano a cero en los cinco individuos (Fig. 7) y la representación de la exactitud (ACC) alcanza el 100 % en por lo menos 3 sujetos (Fig. 8).

| Métrico | SU1 | SU2 | SU3 | SU4 | SU5 | AVG |
|---|---|---|---|---|---|---|
| NRMSE | 0.002092 | 0.000007 | 0.021933 | 0.000003 | 0.060161 | 0.01684 |
| ACC | 1.000000 | 1.000000 | 0.998667 | 1.000000 | 0.998667 | 0.99947 |
| F1S | 1.000000 | 1.000000 | 0.996687 | 1.000000 | 0.996687 | 0.99867 |
| BER | 0.000000 | 0.000000 | 0.001058 | 0.000000 | 0.001058 | 0.00042 |

*Tabla 1. Métricos de desempeño obtenidos en validación*

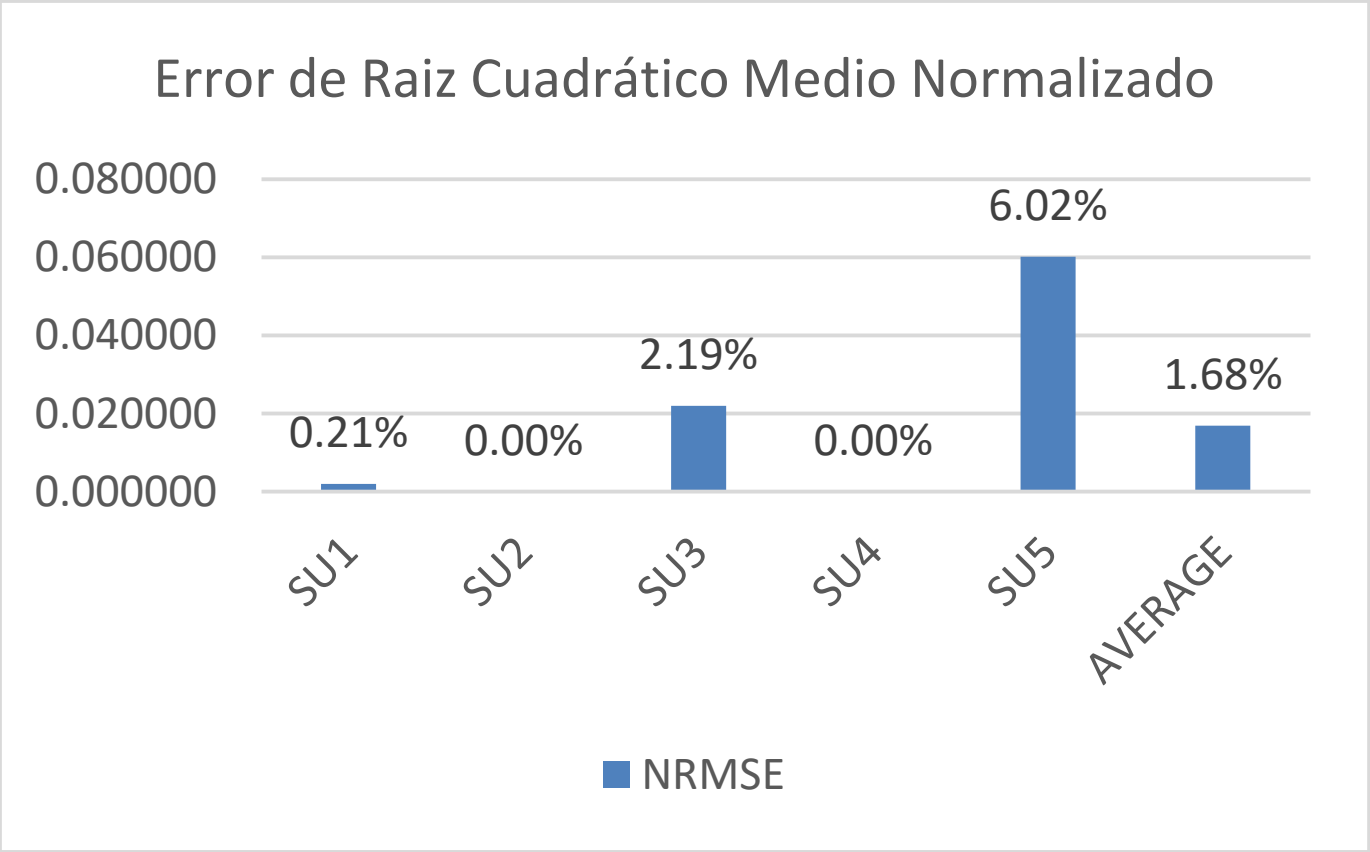

*Fig. 7. Grafica representativa del Error de Raíz Cuadrático Medio Normalizado (NRMSE) en validación.*

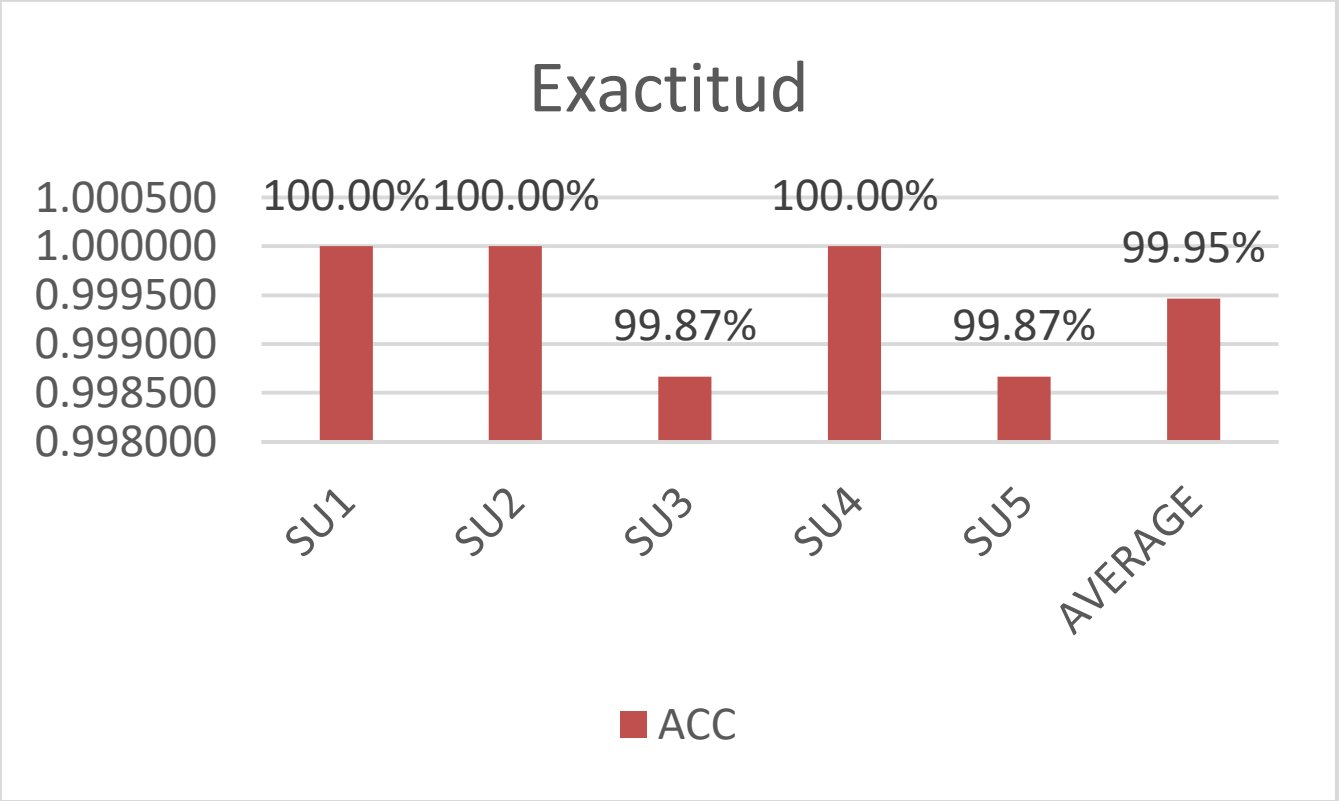

*Fig. 8. Grafica representativa de la exactitud (ACC) en validación.*

En la Fig. 9, se observa el promedio ponderado de la *precisión* y *la exhaustividad*. Lo que significa que esta puntuación toma en cuenta tanto falsos positivos como falsos negativos. F1, es por lo general más útil que la precisión, en especial si se tiene una distribución de clases diferentes [19].

.

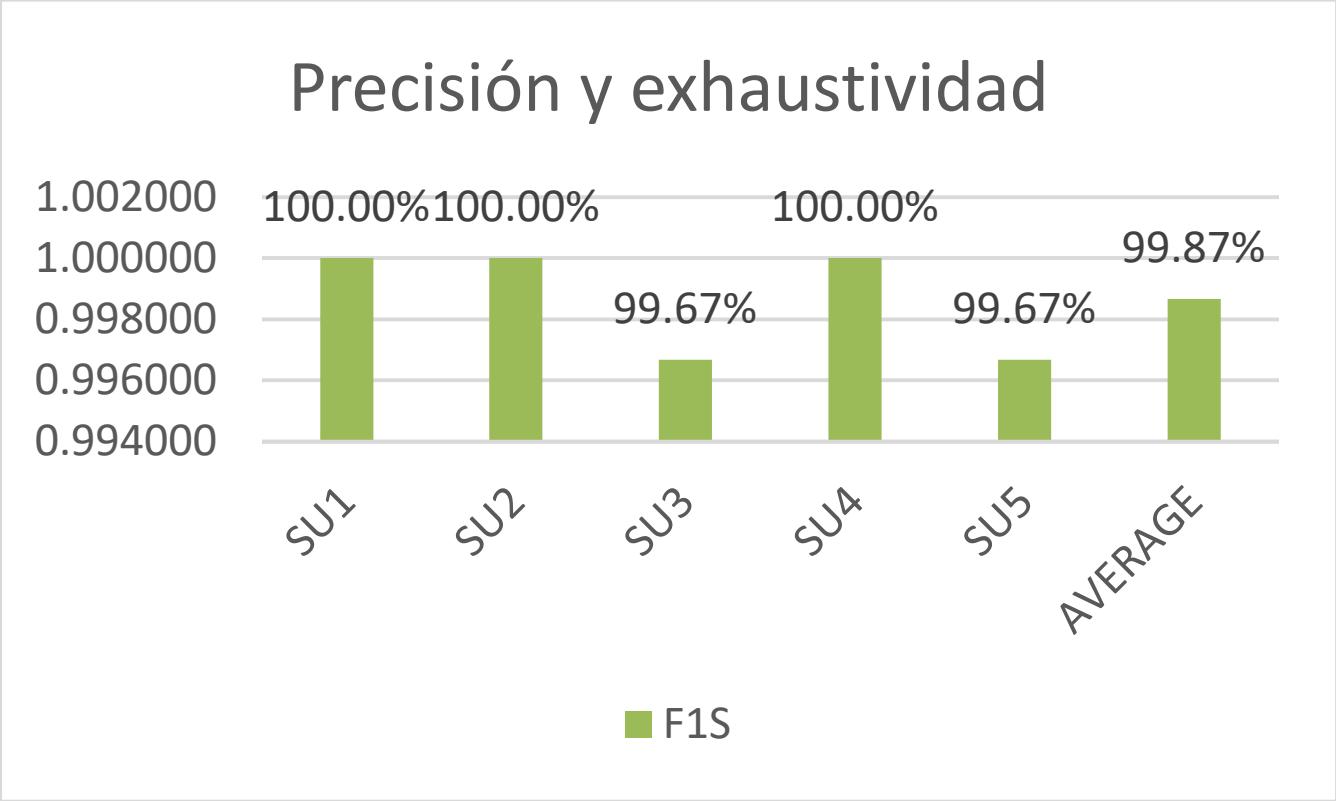

*Fig. 9. Grafica representativa de medicion de la exactitud (F1S) en validación.*

En la Fig.10, se puede observar como la tasa de error balanceada es ligeramente mayor en los inviduos 3 y 5. Esta herramienta de desempeño es la media de los errores de cada clase [20].

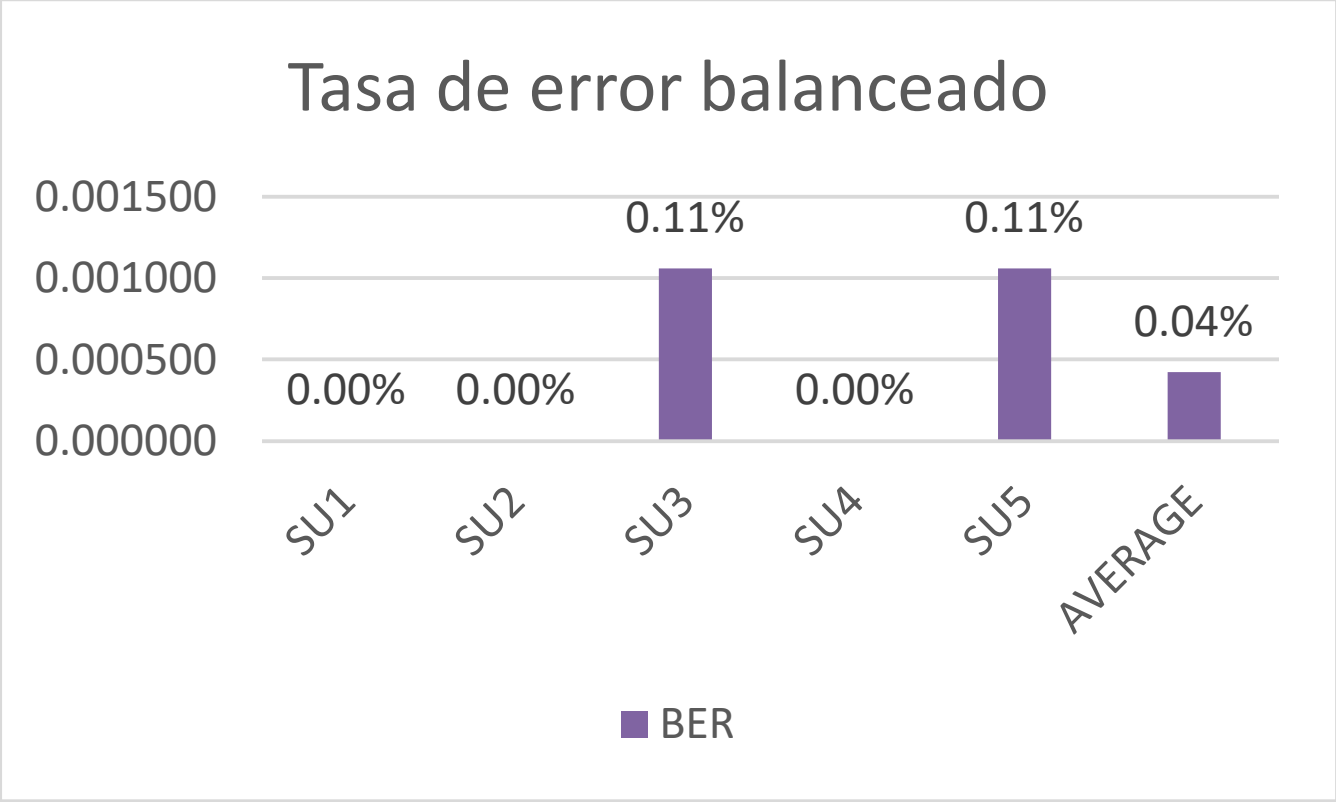

*Fig. 10. Grafica representativa de la tasa de error balanceada (BER) en validación.*

## 5. CONCLUSIONES

En base a los resultados anteriores se puede determinar que en promedio la red tiene una exactitud del 99.94% y un 0.01684 de error para resolver el problema del reconocimiento de 5 imágenes de profundidad del lenguaje de señas americano. Como trabajo futuro se propone realizar el entrenamiento de la RN con imágenes con imágenes que contengan diferente posición y ángulo, ocasionando una imagen con cierta distorsión en sus pixeles, esto con el fin de, tener un sistema más robusto.

## 6. REFERENCIAS